\def\year{2022}\relax
\documentclass[letterpaper]{article} 
\usepackage{aaai22}  
\usepackage{times}  
\usepackage{helvet}  
\usepackage{courier}  
\usepackage[hyphens]{url}  
\usepackage{graphicx} 
\usepackage{natbib}  
\usepackage{caption} 
\DeclareCaptionStyle{ruled}{labelfont=normalfont,labelsep=colon,strut=off} 
\usepackage{algorithm}
\usepackage{algorithmic,amsmath}
\usepackage{amsthm,amssymb}
\usepackage{hyperref}
\theoremstyle{definition}
\newtheorem{definition}{Definition}[section]
\newtheorem{theorem}{Theorem}

\usepackage{newfloat}
\usepackage{listings}
\floatstyle{ruled}
\newfloat{listing}{tb}{lst}{}
\floatname{listing}{Listing}
\title{Exploring the Unfairness of DP-SGD Across Settings}
\author{Frederik Noe, Rasmus Herskind and Anders S{\o}gaard\\
University of Copenhagen}
\usepackage{bibentry}

\begin{document}

\maketitle

\begin{abstract}
End users and regulators require private and fair artificial intelligence models, but previous work suggests these objectives may be at odds. We use the CivilComments to evaluate the impact of applying the {\em de facto} standard approach to privacy, DP-SGD, across several fairness metrics. We evaluate three implementations of DP-SGD: for dimensionality reduction (PCA), linear classification (logistic regression), and robust deep learning (Group-DRO). We establish a negative, logarithmic correlation between privacy and fairness in the case of linear classification and robust deep learning. DP-SGD had no significant impact on fairness for PCA, but upon inspection, also did not seem to lead to private representations.
\end{abstract}


\section{Introduction}

The vast majority of artificial intelligence research projects focus on learning accurate models, but in practice, real-life application of machine learning models require balancing performance with the need to prevent discrimination against protected demographic subgroups and satisfying privacy principles. Both needs are prerequisites for employing machine learning models at scale in many domains, and the lack of fairness and privacy in many widely used models threatens the public perception of artificial intelligence. Unfortunately, the needs are seemingly at odds \citep{pmlr-v81-ekstrand18a,cummings2019compatibility,bagdasaryan2019differential,farrand2020neither,chang2021privacy,agarwal21tradeoffs}. Common fairness and privacy objectives are in conflict, because privacy-preserving algorithms often disproportionally affect members of minority classes \citep{farrand2020neither}.\footnote{Note this is a different trade-off than the fairness-privacy trade-off which results from the need for collecting sensitive data to learn fair models \citep{veale2017fairer}.} The exact nature of this trade-off has not been analyzed, however, and perhaps more surprisingly,  
there has been no evaluations of whether learning algorithms that were developed with group fairness in mind, such as \citet{sagawa2020distributionally}, are less sensitive to optimization with DP-SGD.  

\begin{figure}
    \centering
    \includegraphics[width=0.99\columnwidth]{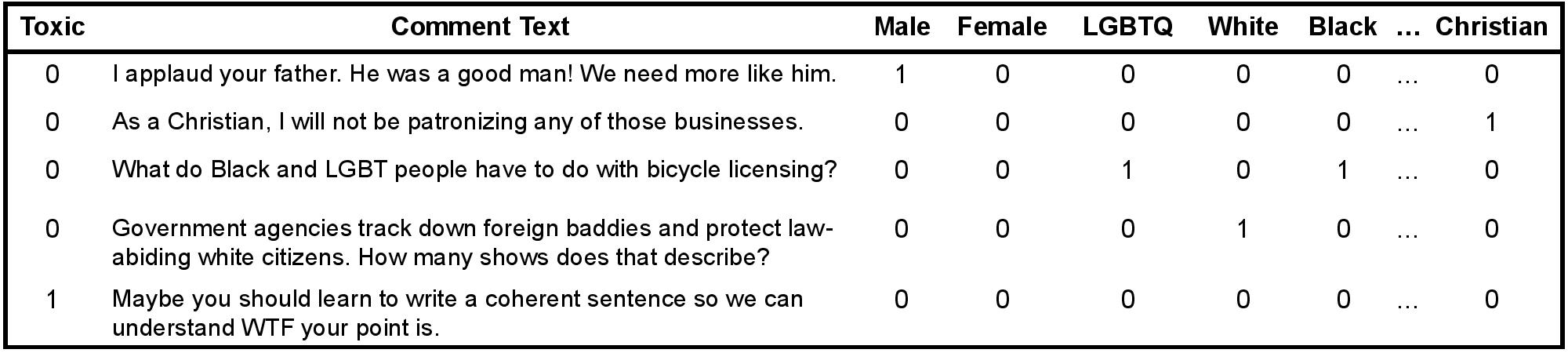}
    \caption{A few datapoints from CivilComments including demographic identities ~\cite{koh2021wilds}}
    \label{fig:dataset}
\end{figure}

\paragraph{Contributions} We evaluate the impact on fairness (Unequal Risk) of using DP-SGD optimization, across three settings: PCA, logistic regression and Group-DRO \cite{sagawa2020distributionally}. For the baseline condition, we confirm past results and establish a negative correlation between fairness and privacy. We further show that this negative correlation is {\em logarithmic}. Moreover, we show that DP-SGD has little impact on fairness in the PCA setting. This, however, seems to because differentially private dimensionality reduction is insufficient to guarantee the privacy of downstream classifiers. 

\section{Fairness and Privacy}

We represent each data point
as $z = (x, g, y) \in \mathcal{X} \times \mathcal{G} \times \mathcal{Y}$, with $g\in \mathcal{G}$ encoding its protected attributes (e.g., race, gender). Let $\mathcal{D}^g_y$ denote the distribution of data with protected attribute $g$ and label $y$. Several definitions of fairness exist \citep{pmlr-v97-williamson19a}, but we use (generalized) approximately constant conditional (equal) risk \citep{donini18empirical}:

\begin{definition}[$\delta$-Unequal Risk] Let $\ell^{g_i}(\theta)=\mathbb{E}[\ell(\theta(x), y)|g=g_i]$ be the risk of the samples in
the group defined by $g_i$, and $\delta\in [0, 1]$. We say that a model $\theta$ exhibits $\delta$-Unequal Risk if for any two values of $g$, say $g_i$ and $g_j$,  $|\ell^{g_i}(\theta)-\ell^{g_j}(\theta)|<\delta$.
\end{definition}

Note that if $\ell$ coincides with the performance metric of a task, and $\delta=0$, this is identical to performance or classification parity \citep{yuan2021assessing}.\footnote{Such a notion of fairness can
be derived from John Rawls’ theory on
distributive justice and stability, treating model performance as a resource to be allocated. Rawls' {\em difference principle}, maximizing the welfare
of the worst-off group, 
is argued to lead to stability and mobility in society at large \citep{rawls_theory_1971}. Performance or classification parity has, however, been argued to suffer from statistical limitations in \cite{corbettdavies2018measure}, which remind us that when risk distributions
differ, standard error metrics are poor proxies of individual equity. This is known as the problem of infra-marginality. Note, however, that this argument does not apply to binary classification problems.}  $\delta$ directly measures what is sometimes called Rawlsian {\em min-max fairness} \citep{10.1287/opre.1100.0865}.  
Recall the standard definition of  $(\varepsilon,\phi)$-privacy\footnote{We talk about $(\varepsilon,\phi)$-privacy rather than $(\varepsilon,\delta)$-privacy, to avoid confusion with $\delta$-fairness.} is as follows: 

\begin{definition}[$(\varepsilon,\phi)$-privacy] $\theta$ is $(\varepsilon,\phi)$-private iff 
    $\mbox{Pr}[\theta(\mathcal{X})]\leq \exp(\varepsilon)\times  \mbox{Pr}[\theta(\mathcal{X}')]+\phi$
for any two distributions, $\mathcal{X}$ and $\mathcal{X'}$, different at most in one row.\end{definition}

\noindent Differential privacy ensures that an algorithm will generate similar outputs
for similar training data sets. Note the multiplicative bound $\exp(\varepsilon)$ and the additive bound $\phi$ serve different roles: The $\phi$ term represents the possibility that a few data points are not governed by the multiplicative bound, which controls the level of privacy (rather than its scope). Note that it also follows directly that if $\varepsilon=0$ and $\phi=0$, absolute privacy is required, leading $\theta$ to be independent of the data. In our experiments, we assume $\phi=0$ and look at the impact of DP-SGD training across different values of $\epsilon$. 


\citet{agarwal21tradeoffs} shows that a $(\varepsilon,0)$-private and fully fair model -- using equalized odds as the definition of fairness -- will be unable to learn anything. To see this, remember that a fully private model is independent of the data and unable to learn from correlations between input and output. If $\theta$ is, in addition, required to be fair, it is thereby required to be fair for all distributions, which prevents $\theta$ from encoding any prior beliefs about the output distribution. Equal odds is often less interesting for NLP applications than equal risk, but note this finding generalizes straight-forwardly to equalized risk, and to approximate fairness (since even for finite distributions, we can define a $\delta>0$, such that preserving absolute privacy leads to a constant $\theta$). 

\begin{theorem}
For sufficiently small values of $\varepsilon$, a fully $(\varepsilon,0)$-private model $\theta$ that is also $\delta$-fair,  will have trivial performance. 
\end{theorem}

\begin{proof}
This follows directly from the above. 
\end{proof}

\noindent We focus on $\delta$-Unequal Risk below, but for completeness include alternative definitions. While $\delta$-Unequal Risk only considers the difference between extremes, $\delta$-Variance considers the overall F$_1$-score variance across the groups, with $\delta=\sigma^2$ defined as the mean of the squared distances of $\ell^{g_i}$ from their mean $\Bar{\ell^{g_i}}$, i.e.., $\frac{\sum_{i=1}^n (\ell^{g_i} - \Bar{\ell^{g_i}})^2}{n}$. Often Unequal Risk is contrasted with Equal Odds, i.e., the idea that fairness requires similar positive rates across groups. \citet{Zafar} introduce the so-called {\em p\%-rule}, a variant of Equal Odds, which we will also adopt below. In the binary case, we say a model $\theta$ exhibits $p$-Equal Odds if 
$$p=\min_{i,j}\min(\frac{\mathbb{E}[\theta(\mathcal{X}^{g_i})]}{\mathbb{E}[\theta(\mathcal{X}^{g_j})]},\frac{\mathbb{E}[\theta(\mathcal{X}^{g_j})]}{\mathbb{E}[\theta(\mathcal{X}^{g_i})]})$$

\noindent Note that if the p\%-rule returns 1.0, $\theta$ exhibits 1.0-Equal Odds, which amounts to identical positive rates across groups. It is easy to show that $p\in[0,1]$, with higher values suggesting fairness under Equal Odds. We modify the p\%-rule, which serves as a compromise between Unequal Risk and Equal Odds. 

\begin{definition}[Modified p\%-rule] We say a model $\theta$ exhibits $p$-Equally Correct Priors if 
$$p=\min_{i,j}\min(
    \frac{
        \frac{\mathbb{E}[\theta(\mathcal{X}^{g_i})]}
            {\mathbb{E}[\mathcal{Y}^{g_i}]}}
            {
        \frac{\mathbb{E}[\theta(\mathcal{X}^{g_j})]}
            {\mathbb{E}[\mathcal{Y}^{g_j}]}
            },
    \frac{
        \frac{\mathbb{E}[\theta(\mathcal{X}^{g_j})]}
            {\mathbb{E}[\mathcal{Y}^{g_j}]}}{
        \frac{\mathbb{E}[\theta(\mathcal{X}^{g_i})]}
            {\mathbb{E}[\mathcal{Y}^{g_i}]}
            }
    )$$
\end{definition} 

\noindent This definition says the divergence from the positive class prior should be no different across any two groups. Note how this, unlike $\delta$-Unequal Risk, says {\em nothing}~about the actual risk of misclassification. It is merely a weighted version of $p$-Equal Odds, which takes the true priors into account. We report results for $\delta$-Variance and the p\%-rule in the Appendix, and focus on $\delta$-Unequal Risk and the Modified p\%-rule here. 

\begin{figure}
    \centering
    \includegraphics[width=3in]{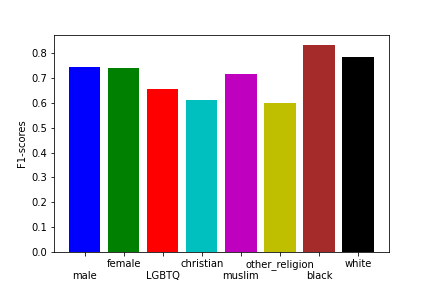}
    \caption{Baseline performance (F$_1$) across groups}
    \label{fig:baseline}
\end{figure}

\section{Experiments}

Our experiments evaluate the impact of differential privacy \citep{10.1007/11681878_14} on the fairness of different machine learning architectures. Differentially private stochastic gradient descent (DP-SGD) is the {\em de facto} baseline approach to learning private models, and we limit ourselves to variants of DP-SGD. Vanilla DP-SGD limits the influence of training samples by (i) clipping the per-batch gradient where its norm exceeds a pre-determined clipping bound $C$, and by (ii) adding Gaussian noise $\mathcal{N}$ characterized by a noise scale $\sigma$ to the aggregated per-sample gradients. We control this influence with a privacy budget $\varepsilon$, where lower values for $\varepsilon$ indicates a more strict level of privacy. DP-SGD has remained popular, among other things because it generalizes to iterative
training procedures \citep{48293}, and supports tighter bounds using the Rényi method \citep{8049725}. We evaluate the impact of differentially private training on fairness across dimensionality reduction, linear classification, and deep learning. 

\paragraph{Bertweet and DistilBert} Transformer architectures are often trained with masked language modeling objectives to learn representations that are useful for downstream NLP tasks \citep{devlin-etal-2019-bert,liu2019roberta}. We use two pretrained, Transformer-based language models: (a) Bertweet \citep{nguyen2020bertweet} is trained on English tweets with the same hyper-parameters as the original BERT-base model \cite{devlin2019bert}. DistilBert \cite{sanh2020distilbert} is induced by distillation of BERT \cite{devlin-etal-2019-bert} to obtain a more efficient and smaller version of it. 

\paragraph{Principal Component Analysis} 
Principal components are linear combinations or mixtures of the initial variables and internally uncorrelated to minimize the information loss. We rely on the IBM Differential Privacy Library (Diffprivlib)\footnote{\url{https://github.com/IBM/differential-privacy-library}} \cite{diffprivlib} for DP-PCA, which relies on \citet{imtiaz16}, who add Wishart noise to this reconstruction to obtain $\varepsilon$-privacy. Their algorithm is shown to approximate the underlying data well, especially for $\varepsilon\geq 1$. Alternatives are discussed in \citet{shang21}. We apply DP-PCA to Bertweet representations of CivilComments and pass these to a three-layer feed-forward network. We varied $\varepsilon$-values in [0.1,10] with a step-size of 0.5, and performed grid search for the hyper-parameters of the feed-forward network. Its dimensionality was given (768, 64, and 32 in the outermost layer), but all other parameters were optimized for validation performance: We trained for 2979 epochs, with learning rate $1e^{-06}$, MSE loss, AdamW optimization, and a batch size of 32. 

\paragraph{Logistic Regression} 
We also evaluate the impact of DP-SGD on linear classification. 
In logistic regression, the weighted sum of model parameters and input is transformed to probabilities by the logistic function, with parameters fit to minimize logistic loss with DP-SGD. We train 400 models with $\varepsilon\in\{0.1, 0.2, 0.3 · · · , 39.9, 40\}$. Again, we rely on Bertweet and Diffprivlib.  

\paragraph{Group Distributionally Robust Optimization} We also consider a deep architecture 
optimized for group fairness, namely GroupDRO \citep{sagawa2020distributionally}, which relies on distributionally robust optimization \citep{hashimoto2018fairness} to minimize the worst-case loss over the groups in the training data. 
We use DP-SGD in Opacus\footnote{\url{https://github.com/pytorch/opacus}} \cite{opacusAPI} to implement differentially private GroupDRO. DP-SGD was generalized to deep learning in \citet{abadi2016deep}; Opacus uses Renyi Differential Privacy (RDP) \cite{Mironov_2017}.

\paragraph{Data} We use Civil Comments \cite{koh2021wilds}, a toxicity detection dataset in which comments are annotated for toxicity and {\em the demographics of the target of toxicity} (see Fig.~\ref{fig:dataset}). 
The task is a binary classification task. 
There are eight target groups:\footnote{We cite the names used in the publicly available dataset. It is unclear to what extent the names impacted annotations. Some seem to represent communities that can in part be referred to by other, more general or more specific, terms, e.g., LGBT, LGBTQ+, the Rainbow Community. By using this data, we do not mean to imply that these terms are useful for referring to groups of citizens. We use CivilComments simply because it is the most widely distributed NLP dataset for fairness studies.} LGBTQ, male, female, Christian, Muslim, other religions, Black, and White.  For simplicity, we only use data from these groups: 
Prior work has shown that toxicity classifiers pick up on biases in the training data and spuriously associate toxicity with the mention of certain demographics ~\cite{civilcomments}. Thus, for classifying the toxicity of online comments, the setup shown in Figure \ref{fig:dataset} ensures that we are able to keep track of fairness by analysing scores across different groups.

\begin{figure}[h]
    \centering
    \includegraphics[width=2.4in]{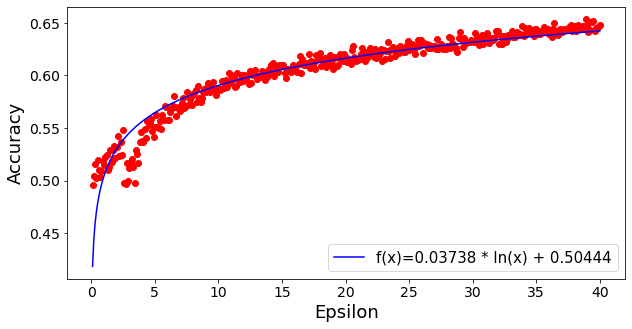}
    \includegraphics[width=1.1in]{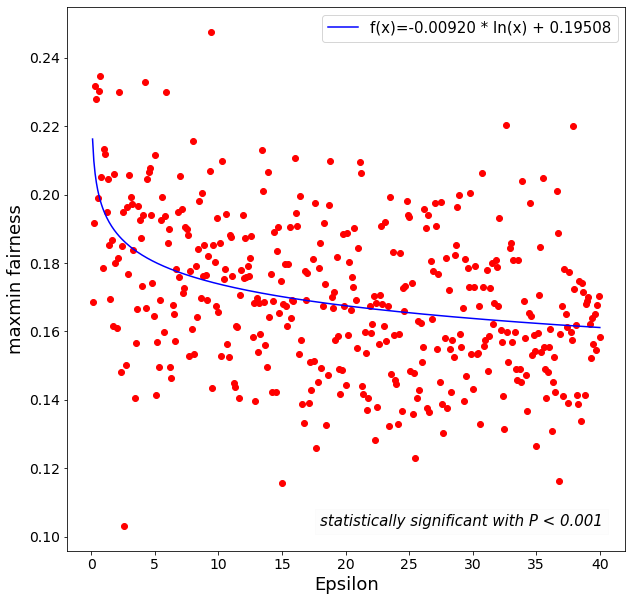}
    \includegraphics[width=1.1in]{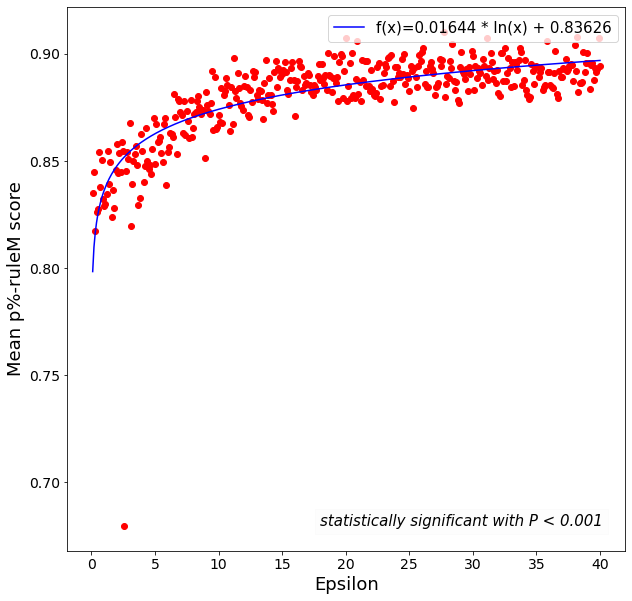}
    \includegraphics[width=2.4in]{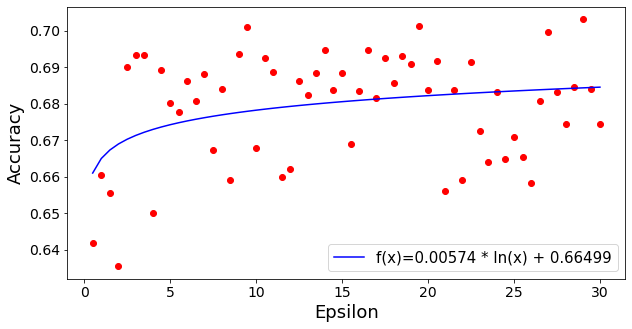}
    \includegraphics[width=1.1in]{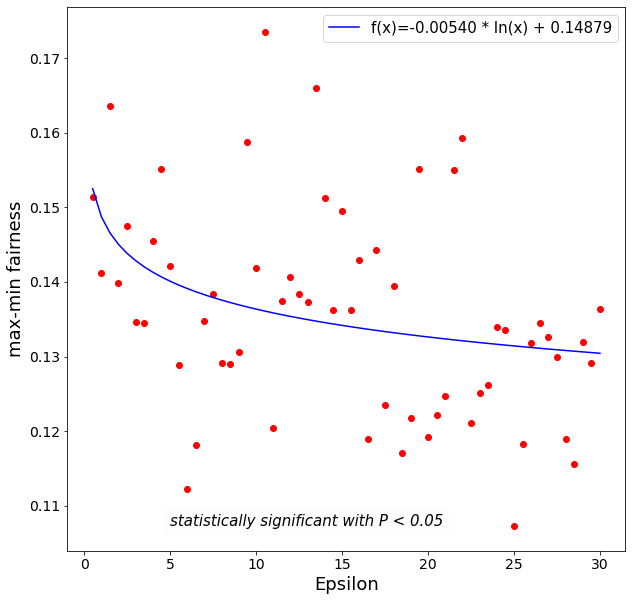}
    \includegraphics[width=1.1in]{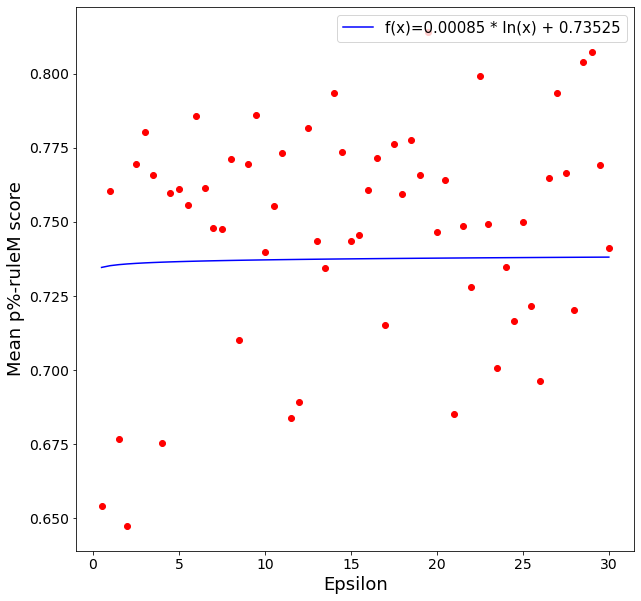}
    \caption{\label{fig:logregr-equal}{\bf Top} ({\em Linear Classification}): Logarithmic correlations between $\varepsilon$ ($x$) and accuracy ($y$) ({\bf first} row), $\varepsilon$ and Unequal Risk ({\bf second}, left), and  $\varepsilon$ and the modified p\%-rule ({\bf second}, right). {\bf Bottom} ({\em Deep GroupDRO}): Logarithmic correlation between $\varepsilon$ ($x$) and accuracy ($y$) ({\bf third}), $\varepsilon$ and Unequal Risk ({\bf fourth}, left), and the modified p\%-rule ({\bf fourth}, right).}
\end{figure}

\section{Results}


The baseline F$_1$ scores across the eight groups are plotted in Figure~\ref{fig:baseline}. Scores significantly across the eight groups, with lowest toxicity detection scores for minority religious groups and Christians, and highest toxicity detection scores for blacks and whites, reportedly being the prototypical targets for American and European racist slur \cite{Croom2015-CROSSA}. Generally, PCA reduces performance slightly, while our GroupDRO architecture performs slightly better than our baseline with fixed Bertweet embeddings. We are mainly interested in how overall performance and fairness is affected by DP-SGD. 

Figure~\ref{fig:logregr-equal} ({\bf second row}) shows the correlation between the degree of approximate privacy and Unequal Risk for linear classification: The more private a model, the less fair. This correlation contrasts with the positive correlation between $\varepsilon$ and accuracy ({\bf first}): The more private a model, the worse it performs. Note that both of these correlations are {\em significant} ($p<0.001$) and {\em logarithmic} and have similar coefficients. 



GroupDRO with DistilBert exhibits similar correlations; see Figure~\ref{fig:logregr-equal}. 
While odds become logarithmically more equal with higher $\varepsilon$ values for linear models, odds exhibit high variance for GroupDRO and are not significantly correlated with $\varepsilon$. The fact that GroupDRO exhibit a similar, negative, logarithmic correlation between $\epsilon$ and Unequal Risk, is particularly interesting, since GroupDRO assumes access to group information during training and uses this information to learn a model with Equal Risk. Under sufficient privacy guarantees, the approach is ineffective, however. 

In contrast, with PCA, we see a weak, linear correlation between $\varepsilon$ and performance: The less privacy, the better performance, again, but differences are minimal (absolute drops in accuracy within a 0.5 margin). We also see a weak, negative, linear correlation between $\varepsilon$ and Unequal Risk: The less privacy, the more fairness, but this correlation is {\em not}~significant. At first sight, it may seem promising that differentially private dimensionality reduction does not lead to Unequal Risk. The same pattern is seen across all the fairness measures. However, this is unfortunately {\em not} a case of getting the best of both worlds: Our feed-forward network seems to be able to recover private information from our dimensionality-reduced representations. To see why this may happen, consider that the algorithm presented in \citet{imtiaz16} is only approximate and not providing strict guarantees. It adds stochastic noise to each dimension independently, and only provides expected guarantees. The more dimensions that carry signal, the higher the chance that private information is recoverable. The sensitivity of differentially private dimensionality reduction has been observed before \cite{Aggarwal05}. We hypothesize that our feed-forward network in a way similar to spectral filtering \cite{kargupta03} can filter off the random noise by implicitly analyzing eigenstates. 

\section{Discussion and Conclusion}

The trade-off between fairness and privacy is of great concern to the machine learning community, but previous work has not suggested promising research directions for how to best solve this dillemma. Our results go beyond previous work on the trade-offs between fairness and privacy \citep{bagdasaryan2019differential,agarwal21tradeoffs} in evaluating this trade-off across multiple machine learning settings and in establishing {\em logarithmic}~correlations between fairness and privacy across two different settings: linear classification and robust deep learning. Finally, we also provide an example of a successful {\em attack} against differentially private dimensionality reduction by training a three-layer feed-forward network on the learned representations of DP-PCA. 



\citet{uniyal2021dpsgd} presented a proposal, which exhibited a better trade-off, empirically. \citet{liu2021fair} presented an approach to fairness and privacy that relies on a self-adaptive mechanism and dynamically adjusts instance influence in each class depending on the theoretical bias-variance bounds.\footnote{The paper was rejected, though, because it did not prove reasonable privacy guarantees, and because it relied on non-standard definitions of fairness. See \url{https://openreview.net/forum?id=IqVB8e0DlUd}} \citet{fioretto2021decision} proposes various solutions, including output perturbations, linearizations, learning piece-wise linear proxy-functions, or fairness payment. These methods have several drawbacks, and most are not generally applicable. Piece-wise linear proxy-functions and GroupDRO assume that demographics are available during training, which limits their application severely. In summary, finding robustly applicable approaches to simultaneously optimizing for fairness and privacy remains an open problem. Our contributions are (a) showing the {\em logarithmic} nature of this correlation, suggesting the usefulness of maximizing the second order gradient of this trade-off; (b) showing this holds even for group-robust deep learning; and (c) showing, by example, that seemingly good results for {\em both} fairness and privacy may be the result of implicit privacy attacks, when deep classifiers exploit privacy leaks. 


Based on a sequence of experiments with toxicity detection, relying on large-scale pretrained language models, this paper established the logarithmic nature of the relationship between fairness and privacy, in the form of DP-SGD, and showed that this logarithmic nature generalizes to models trained to optimize for fairness (with GroupDRO). Moreover, we saw that when applied to dimensionality reduction, DP-SGD did not seem to violate fairness. This anomaly was explained by leakage of private information, however. We discussed possible research directions for jointly learning private and fair machine learning models, a topic we believe to be of great importance to the future of artificial intelligence. 

\bibliography{bib}
\bibliographystyle{acl_natbib}

\end{document}